\documentclass{Interspeech}

\interspeechcameraready

\title{Phir Hera Fairy: An English Fairytaler is a Strong Faker of Fluent Speech in Low-Resource Indian Languages}

\author[affiliation={1}]{Praveen}{Srinivasa Varadhan}
\author[affiliation={1}]{Srija}{Anand}
\author[affiliation={2}]{Soma}{Siddharta}
\author[affiliation={1}]{Mitesh}{M. Khapra}

\affiliation{AI4Bharat}{Indian Institute of Technology Madras}{India}
\affiliation{}{Saryps Labs}{India}
\email{cs21d201@cse.iitm.ac.in, miteshk@cse.iitm.ac.in}
\keywords{speech synthesis, zero-resource, fine-tuning.}

\usepackage[table,xcdraw]{xcolor}
\usepackage{comment}
\usepackage{booktabs}
\usepackage{tcolorbox}
\usepackage{multirow}
\usepackage{booktabs, colortbl} 
\usepackage{nicematrix}

\usepackage{hyperref}
\usepackage{xurl}
\hypersetup{
  breaklinks   = true,           
  pdfborder    = {0 0 0},        
  colorlinks   = false           
}

\usepackage[normalem]{ulem}  

\definecolor{customOrangePastel}{HTML}{f9dcc0}
\definecolor{customOrange}{HTML}{FFE0B2}
\definecolor{customPinkPastel}{HTML}{ffe5e7}
\definecolor{customVioletPastel}{HTML}{eadaf5}
\definecolor{customPink}{HTML}{FCE4EC}
\definecolor{customBlue}{HTML}{E3F2FD}
\definecolor{deepBlue}{HTML}{0D47A1}
\definecolor{deepRed}{HTML}{C62828}
\definecolor{deepOrange}{HTML}{D84315}
\definecolor{deepViolet}{HTML}{662094}
\definecolor{customGreen}{HTML}{DCEDC8}

\begin{document}

\maketitle

 \begin{abstract}
What happens when an English Fairytaler\footnote{Reference to English F5-TTS: A Fairytaler that Fakes Fluent and Faithful Speech With Flow Matching} is fine-tuned on Indian languages? 
We evaluate how the English F5-TTS model adapts to 11 Indian languages, measuring polyglot fluency, voice-cloning, style-cloning, and code-mixing. We compare: (i) training from scratch, (ii) fine-tuning English F5 on Indian data, and (iii) fine-tuning on both Indian and English data to prevent forgetting. Fine-tuning with only Indian data proves most effective and the resultant IN-F5\footnote{\url{https://huggingface.co/ai4bharat/IndicF5}}
 is a near-human polyglot; that enables speakers of one language (e.g., Odia) to fluently speak in another (e.g., Hindi). Our results show English pretraining aids low-resource TTS in reaching human parity. To aid progress in other low-resource languages, we study data-constrained setups and arrive at a compute optimal strategy. Finally, we show IN-F5 can synthesize unseen languages like Bhojpuri and Tulu using a human-in-the-loop approach for zero-resource TTS via synthetic data generation.

\end{abstract}

\section{Introduction}

Text-to-Speech (TTS) synthesis has seen remarkable advancements, particularly in English, where state-of-the-art (SOTA) models now approach human parity in naturalness~\cite{DBLP:journals/corr/abs-2409-00750, DBLP:journals/pami/TanCLCZLWLYHZQSL24, anastassiou2024seed}, speaker adaptation~\cite{shen2024naturalspeech, DBLP:conf/acl/Peng00MH24, borsos2023soundstorm}, and expressiveness~\cite{ju2024naturalspeech,le2024voicebox, du2024cosyvoice}. These breakthroughs are largely driven by the scaling of both data and model size. For instance, the F5-TTS~\cite{chen2024f5} model has been trained on 100K hours of speech data, requiring massive computational resources. Modern TTS models are no longer just about monotonic speech synthesis but they exhibit emergent abilities such as human-level natural speech, zero-shot voice cloning, expressive synthesis, and polyglot capabilities.

While these advancements are impressive, they remain largely inaccessible to low-resource languages, where both data and compute constraints pose significant barriers. Existing efforts in multilingual TTS, such as VoiceBox~\cite{le2024voicebox} and YourTTS~\cite{casanova2022yourtts}, have demonstrated that fine-tuning or prompting on as little as few minutes of data can yield intelligible speech synthesis in low resource languages. 
However, most efforts have been constrained to basic text-to-speech synthesis and intelligibility, with limited exploration \cite{Kumar2022TowardsBT} of the emergent abilities  mentioned above. Emergent abilities like polyglot fluency and seamless handling of code-mixed speech are not just desirable features but essential for inclusivity in linguistically diverse countries like India, where bilingualism and multilingualism are the norm rather than the exception.
To drive this inclusion, we ask: How can we accelerate TTS technology for low-resource languages while being mindful of compute and data sparsity?

In this work, we investigate whether large-scale English TTS models which are already equipped with emergent abilities, can serve as a strong prior for low-resource multilingual synthesis. Our goal is to bypass the prohibitive costs of large-scale data collection and compute by leveraging an English-trained checkpoint. To systematically investigate this, we evaluate three key strategies while using only one-hundredth of the total data used to train a state-of-the-art English TTS model, even when summing across 11 languages: (i) training from scratch on 11 Indian languages (IN11), (ii) fine-tuning an English-trained model on IN11, and (iii) fine-tuning on IN11 while retaining English data. Our findings challenge conventional assumptions about low-resource TTS training, 
demonstrating exceptional fluency, natural prosody, polyglot capabilities, speaker adaptation, and seamless code-switching, as measured by MUSHRA~\cite{varadhan2024rethinking} scores.

We further extend the boundaries of conventional adaptation by going beyond the need for even minimal fine-tuning data in new languages. Specifically, we explore zero-resource TTS by demonstrating that a fine-tuned model can synthesize intelligible speech in Bhojpuri and Tulu, despite the absence of direct training data for these languages. This is made possible through transfer learning from linguistically related languages with shared scripts. To further refine this approach, we incorporate human-in-the-loop filtering and synthetic data augmentation, thereby proposing a scalable framework for extending TTS capabilities to entirely unseen languages. Our method achieves MUSHRA scores of 82.0 and 93.6 for Bhojpuri and Tulu, respectively, emphasizing its effectiveness in building inclusive TTS for under-represented languages. This work opens new avenues for leapfrogging traditional barriers in TTS technology, demonstrating that with minimal compute and data, we can achieve high-quality speech synthesis and unlock emergent abilities for underserved languages.

\section{Methodology}

In this section, we describe how we adapt F5-TTS for Indian languages by expanding its vocabulary and evaluating different training strategies for achieving human-like synthesis. To understand effects of scaling data, we ablate fine-tuning on 1, 10, and 100-hour splits per language. Finally, we propose a zero-resource TTS recipe, leveraging IN-F5 to synthesize speech for unseen languages which have no available data.

\subsection{Enabling the Adaptation of F5 to Indian Languages}
To adapt F5-TTS for Indian languages, we first extend its vocabulary by inserting raw characters corresponding to Indian native scripts, eliminating the need for phoneme models. Indian languages exhibit high phonetic orthography, where written characters closely map to pronunciation, making character-based modeling a natural choice~\cite{anand24_interspeech}. This approach avoids the need for grapheme-to-phoneme (G2P) conversion, which remains underdeveloped for many Indian languages. The vocabulary consists of  685 unique character tokens, covering diverse scripts in IN11. To ensure stable fine-tuning and faster convergence, we initialize embeddings from the same latent space of embeddings of the English pretrained checkpoint by randomly sampling embeddings corresponding to existing tokens. 

\subsection{What Is the Best Way to Teach a Fairytaler New Languages?}
Adapting an English-pretrained model using data from Indian languages (IN11) can follow multiple strategies, but the most effective approach for achieving high-quality speech synthesis is not immediately clear. We compare three strategies:

\noindent \textbf{(i) ($\Phi$→IN) Training from Scratch: } The model is initialized randomly and trained solely on IN11 data, serving as a multilingual baseline.

\noindent  \textbf{(ii) (EN→IN) Direct Fine-Tuning on IN11:} The 100K-hour English-pretrained F5 checkpoint is fine-tuned on IN11 to assess whether prior exposure to English benefits Indian language adaptation.

\noindent  \textbf{(iii) (EN→EN+IN) Fine-Tuning on IN11 with English:} The model is fine-tuned on a mixture containing equal amounts of English \cite{he2024emilia} and IN11 data, allowing us to examine whether continued English exposure enhances generalization while improving performance on Indian languages.

\noindent By evaluating these strategies, we aim to identify the most data-efficient and effective approach for achieving human-like synthesis in low-resource Indian languages.

\subsection{Data Scaling Effects while Fine-tuning in Low-Resource Settings}

High-quality studio-recorded datasets 
typically contain around 10 hours of data per language, while low-resource languages often have less than 1 hour. However, past work shows that even for low-resource languages, 100-200 hours of training data can be obtained by restoring low-quality ASR data using speech enhancements and denoising models (e.g., IndicVoices-R \cite{sankar2024indicvoices-r}). 
 Given these alternatives, we investigate how fine-tuning data quantity impacts adaptation quality in Indian languages. We train models with 1 hour, 10 hours, and 100 hours per language for 150K steps, evaluating the effect of data scaling on polyglot fluency, voice cloning, and overall synthesis quality. All experiments use the same architecture and hyperparameters, isolating the impact of data size alone. To quantify scaling effects across emergent behaviours, we measure intelligibility (WER) and speaker similarity (S-SIM) (see Section \ref{subsec:evaluation-metrics}) and supplement it with perceptual (MUSHRA) evaluations of the final fine-tuned models.

\subsection{Building Zero-Resource TTS}

To develop TTS for truly zero-resource languages, we test our approach on Tulu and Bhojpuri, two Indian languages without existing TTS models. Tulu has no prior speech data, while Bhojpuri has some existing resources \cite{iiith2024limmits}. We simulate a zero-resource setting for Bhojpuri to allow us to compare synthesis quality of our model against actual human recordings. We first curate text corpora from translation datasets \cite{narayanan-aepli-2024-tulu} or existing corpora \cite{iiith2024limmits}. Tulu shares its script with Kannada, while Bhojpuri uses Devanagari, the script used for Hindi. Since Kannada and Hindi are part of our training data, IN-F5 has already learned these script representations, making it well-suited for cross-lingual transfer. Using IN-F5's strong cross-lingual generalization, we synthetically generate 1 hour of speech for each of the two languages from the collected text corpora. We select a Kannada speaker and a Maithili speaker from the Rasa dataset to generate Tulu and Bhojpuri, respectively, ensuring natural prosody and expressiveness given the high quality natural and expressive content in Rasa \cite{ai4bharat2024rasa}. Each generated sample is verified by a native speaker for correctness. We then fine-tune IN-F5 separately on the validated dataset for each language (self-training), following the same training procedure and loss function as in prior fine-tuning experiments. This method demonstrates a scalable framework for rapidly developing TTS models for truly zero-resource languages, with minimal human effort.

\section{Experimental Setup}

\subsection{Datasets}
\textbf{IN11.} We compile a diverse and representative speech-text dataset for 11 Indian Languages, totaling 1417 hours across Assamese, Bengali, Gujarati, Hindi, Kannada, Malayalam, Marathi, Odia, Punjabi, Tamil, and Telugu. To ensure high-quality synthesis, we incorporate studio-quality speech from IndicTTS~\cite{baby2016resources}, LIMMITS~\cite{iiith2024limmits}, and Rasa~\cite{ai4bharat2024rasa}, capturing a range of read speech, conversational dialogues, and expressive styles rendered by professional voice artists. To improve speaker diversity, we integrate Google Crowdsourced TTS~\cite{abraham2020crowdsourcing}, enabling IN-F5 to generalize across different voices recorded in controlled environments. Finally, we leverage IndicVoices-R~\cite{sankar2024indicvoices-r}, a large-scale ASR-restored dataset, as a crucial factor in scaling both speaker variety and overall training data, enhancing the model’s robustness to spontaneous and natural speech across diverse linguistic contexts. From this dataset, we hold out 1100 utterances, carefully balanced across seen and unseen speakers, genders, and age groups, creating the \textit{IN11-Test-Set} for standardized evaluation.
This carefully curated train set allows IN-F5 to achieve high-quality, speaker-adaptive, and expressive TTS across a spectrum of Indian languages.
\subsection{Training Details}
We fine-tune F5 on IN11, starting from the English checkpoint pretrained on nearly 100K hours. The model is fine-tuned for up to 150K steps using the AdamW optimizer with a learning rate of $5 \times 10^{-5}$ and a batch size of 30000 frames per GPU, without gradient accumulation. Training is conducted using mixed precision on 32 NVIDIA H100 Tensor Core GPUs. Spectrograms are computed with 100 mel channels, a 24 kHz sampling rate, a 256 hop length, and a 1024 FFT window size. Warm-up updates are set to 48K steps, with gradient clipping at a max norm of 1.0. Model checkpoints are saved every 2000 updates to enable periodic evaluation and monitor training stability.

\subsection{Evaluation Metrics}
\label{subsec:evaluation-metrics}
We evaluate our models using both subjective and objective metrics to comprehensively assess synthesis quality.
For subjective evaluations to assess the naturalness of systems, we use a variant of MUSHRA \cite{varadhan2024rethinking}, with detailed guidelines and no-mentioned-reference, similar to the setup used in NaturalSpeech2 \cite{shen2024naturalspeech}. IN-F5 demonstrates strong performance across several emergent capabilities of large-scale TTS, including zero-shot voice cloning, cross-lingual generalization, and code-mixing. To specifically evaluate these properties, we introduce two additional subjective scales:
\begin{itemize}
    \item \textbf{MUSHRA-S (Speaker Similarity)}: Measures how closely the synthesized voice matches the reference speaker in tone, pitch, and speaking characteristics. Raters score from 100 (perfect match), 80 (mostly similar), 60 (noticeably different with some similarity), 40 (limited resemblance), 20 (clear mismatch), to 0 (no similarity).
    \item \textbf{MUSHRA-I (Intelligibility)}: Assesses how easily polyglot and code-mixed speech can be understood, with 100 (perfect clarity), 80 (mostly clear with minor mispronunciations), 60 (understandable with mispronunciations), 40 (words skips or unclear), 20 (incoherent speech), and 0 (completely unintelligible or silent).
\end{itemize}

\noindent We report the original MUSHRA scores as Nat (Naturalness) and the two new scores as Sim (Speaker Similarity) and Int (Intelligibility). All human evaluations were conducted by 134 listeners, with an average of 12 native speakers per language, each rating atleast 30 utterances. To objectively evaluate, we measure intelligibility by calculating WER of the SOTA IndicConformer~\cite{javed2024indicvoices} ASR model on the \textit{IN11-test-set}, and speaker similarity by calculating cosine similarity between ground-truth and synthesized embeddings, extracted using WavLM\cite{DBLP:journals/jstsp/ChenWCWLCLKYXWZ22}.

\section{Results}
We present our findings on the adaptation of F5 to Indian languages, examining fine-tuning strategy, emergent behaviors, data scaling effects, and performance against prior baselines. 

\subsection{Are English speech foundation models good priors for multilingual adaptation?}

We present the overall MUSHRA scores for the three fine-tuning strategies in Table \ref{tab:vc-poly-overall}. Intuitively, one might expect that fine-tuning with both English and IN11 data (EN→EN+IN) would yield the best results. However, our findings reveal a surprising trend: direct fine-tuning on IN11 alone (EN→IN) achieves the highest overall MUSHRA score of 73.4, surpassing the other strategies. 
It is possible that this setup reduces English performance, but that is not a concern, as we can always use the original checkpoint for English. Our focus is on developing a multilingual model for low-resource Indian languages for which this setup gives the best performance. 
In the simplest setting with seen speakers and unseen texts, the \textit{naturalness} scores indicate that direct fine-tuning (78.0) slightly surpasses human recordings (75.9), suggesting human-level synthesis. However, the train from scratch model underperforms with an overall MUSHRA of 43.2, underscoring a critical challenge; \textit{while English TTS has reached human parity, replicating this success from scratch in low-resource settings remains extremely difficult.} This stark contrast highlights the importance of large-scale pretraining, but more importantly, it demonstrates a compelling alternative: simply initializing from a large pretrained English model and fine-tuning on a small set of IN11 data (1.4\% of EN data) is sufficient to reach human-level synthesis in Indian languages. This approach not only enables data-efficient multilingual adaptation but also brings Indian TTS closer to parity with English, unlocking new possibilities for low-resource speech synthesis. We refer to the best-performing checkpoint obtained via direct fine-tuning as \textbf{IN-F5} in all subsequent discussions.

\begin{table}[!t]
\centering
\fontsize{8pt}{9pt}\selectfont
\setlength{\tabcolsep}{2pt}
\caption{Comparison of different training strategies for multilingual adaptation to IN11 using MUSHRA. Nat.: Naturalness, Int.: Intelligibility, Sim.: Speaker Similarity. 95\% Confidence intervals: min. = 0.8, avg. = 2.0, max. = 3.2.}
\label{tab:vc-poly-overall}
\begin{tabular}{@{}lccccccccc@{}}
\toprule
\multirow{3}{*}{\textbf{\begin{tabular}[c]{@{}l@{}} \\ Setup\end{tabular}}} & \multicolumn{4}{c}{\cellcolor{customOrange}\textbf{Voice Cloning}}                              & \multicolumn{4}{c}{\cellcolor{customBlue}\textbf{Polyglot}}                                      & \textbf{Overall}     \\ \cmidrule(l){2-5} \cmidrule(l){6-9} \cmidrule(l){10-10} 
& \multicolumn{2}{c}{ \textbf{Seen}} & \multicolumn{2}{c}{ \textbf{Unseen}} & \multicolumn{2}{c}{\textbf{Natural}} & \multicolumn{2}{c}{\textbf{Studio}} & \multicolumn{1}{l}{} \\
& \textbf{\textit{Nat.}}  & \textbf{\textit{Sim.}} & \textbf{\textit{Nat.}}   & \textbf{\textit{Sim.}}  & \textbf{\textit{Nat.}}     & \textbf{\textit{Int.}}    & \textbf{\textit{Nat.}}    & \textbf{\textit{Int.}}    & \multicolumn{1}{l}{} \\ \midrule
Human                                                                                   & 75.9 & 84.6 & 74.6 & 83.2 & \textbf{70.6} & \textbf{74.9} & \textbf{89.7} & \textbf{93.0} & 77.4  \\ 
\arrayrulecolor{gray} 
    \specialrule{0.4pt}{1pt}{2pt} 
    \arrayrulecolor{black} 

\textbf{EN→IN}                                               & \textbf{78.0}           & \textbf{86.8}             & \textbf{76.6}            & \textbf{85.8}              & 68.2              & 72.3             & 77.6             & 83.9             & 73.4                 \\
EN→EN+IN                                                                               & 69.7           & 84.3             & 69.7            & 84.7              & 61.0              & 67.6             & 74.4             & 81.6             & 66.2                 \\
$\Phi$→IN                                                                              & 40.0           & 76.2             & 42.0            & 78.1              & 41.8              & 50.7             & 54.9             & 65.1             & 43.2                 \\ \bottomrule
\end{tabular}
\end{table}

\subsection{Emergent abilities of IN-F5}
We extend our observations on the subjective evaluations presented in Table \ref{tab:vc-poly-overall}, highlighting key emergent abilities of IN-F5.

\noindent \textbf{\colorbox{customOrange}{(i) Voice Cloning.}} 
One of the most notable features of IN-F5 is its advanced voice cloning, achieving higher perceived naturalness and speaker similarity scores compared to human recordings. This enables high-quality, personalized TTS applications for Indian languages. The MUSHRA evaluations in Table \ref{tab:vc-poly-overall} highlight this advantage. For seen speakers, IN-F5 achieves \textit{naturalness} of 78.0, and \textit{speaker similarity} of 86.8, surpassing human recordings at 75.9 and 84.6, respectively. Even for unseen speakers, IN-F5 maintains high performance with \textit{naturalness} at 76.6 and \textit{similarity} at 85.8, again outperforming human recordings (74.6 and 83.2). 
While the higher scores may seem counterintuitive, we attribute the higher speaker similarity scores to the model’s controlled synthesis process, which eliminates the natural inconsistencies in speech, unlike real recordings, which may contain subtle background noise, breath sounds, or microphone artifacts. We find that IN-F5 benefits from the English pre-training and generates clean, noise-free speech, making it sound more natural than human recordings.


\noindent \textbf{\colorbox{customBlue}{(ii) Polyglot.}} More intriguingly, IN-F5 emerges as an exceptional polyglot. In our evaluations, we find that it enables speakers from one linguistic family (e.g., Indo-Aryan Hindi) to fluently generate speech in an entirely different linguistic family (e.g., Dravidian Tamil). To systematically assess IN-F5’s polyglot capabilities, we use native speaker voice prompts from each of the 11 languages to generate speech in the other 10 languages. The model’s outputs, mimicking non-native speakers, are then compared to native human recordings using MUSHRA evaluations. Table \ref{tab:vc-poly-overall} shows that IN-F5 achieves \textit{naturalness} of 68.2, closely tracking the human score of 70.6 for voices in natural environments. However, while IN-F5 excels in such settings, a quality gap remains when compared against studio-quality speech, where humans score near 90 in naturalness and intelligibility, whereas IN-F5 scores 77.6 and 83.9, respectively. Although there is room for improvement, it is important to highlight that such strong cross-lingual cross-speaker generalization has never been demonstrated before for Indian speech synthesis.

\begin{table}[!t]
\centering
\fontsize{8.5pt}{9pt}\selectfont
\setlength{\tabcolsep}{6.5pt}
\caption{Average intelligibility scores (MUSHRA-I) evaluating IN-F5’s code-mixing performance, where each primary language is mixed with 10 target languages (X).}
\label{tab:codemixed}
\begin{tabular}{@{}lcccccc@{}}
\toprule
\textbf{System} & {\cellcolor{customPink}\textbf{IN11-X}} & \textbf{hi-X} & \textbf{ta-X} & \textbf{pa-te} & \textbf{as-te} \\ \midrule
Human           & 81.50               & 88.66         & 79.54         & 80.09          & 72.14          \\
IN-F5           & 76.68               & 85.5          & 76.67         & 79.53          & 52.5           \\ \bottomrule
\end{tabular}
\end{table}

\noindent \textbf{\colorbox{customPink}{(iii) Code-Mixed.}} Table \ref{tab:codemixed} presents intelligibility scores to analyze IN-F5’s code-mixing performance on 30 manually curated sentences from IndicVoices \cite{javed2024indicvoices}. 
We see that IN-F5 shows strong code-mixing ability too, an essential feature for Indian speakers who are inherently multilingual. It fluently synthesizes speech in common code-mixed pairs like Hindi-Bengali and Kannada-Telugu achieving \textit{intelligibility} of 79.8 and {73.1, respectively. Remarkably, it can even generate speech in less natural code-mixed pairs such as for Punjabi-Telugu, scoring a near-human intelligibility score of 79.5 as shown in Table \ref{tab:codemixed}.  While performance drops for more challenging combinations, such as Assamese-Telugu (52.5), this remains a significant milestone, as even human speakers may struggle with such pairings. Moreover, IN-F5 performs exceptionally well in highly spoken languages such as Hindi (hi-X: 85.5), and Tamil (ta-X: 76.7), showing \textit{intelligibility scores} that are comparable to even human recordings in non-code-mixed settings. \\
\noindent \textbf{\colorbox{customGreen}{(iv) Expressive Synthesis.}} We measure IN-F5's ability to render expressive speech across 6 styles - happy, sad, anger, surprise, news-reading, and book narration by providing it appropriate style prompts from the Rasa dataset.  The MUSHRA scores in Table \ref{tab:expressive} reveal that IN-F5 is an ``Excellent'' expressive TTS, that marginally falls behind human recordings.

\begin{table}[]
\centering
\fontsize{9pt}{9pt}\selectfont
\setlength{\tabcolsep}{2pt}
\caption{MUSHRA scores reflecting expressivity of IN-F5 on Rasa test set (Averaged across 8 Indian languages).}
\label{tab:expressive}
\begin{tabular}{@{}ccccccc@{}}
\toprule
\multicolumn{1}{l}{\textbf{}} & \textbf{Happy} & \textbf{Sad}   & \textbf{Anger} & \textbf{Surprise} & \textbf{News}  & \textbf{Book}  \\ \midrule
{\cellcolor{customGreen}IN-F5}                         & 82.8          & 84.8          & 85.4          & 83.4             & 82.3          & 85.2          \\
Human                         & \textbf{93.0} & \textbf{92.3} & \textbf{92.5} & \textbf{92.5}    & \textbf{92.8} & \textbf{91.1} \\ \bottomrule
\end{tabular}
\end{table}



\subsection{Scaling Effects in Low-Resource Adaptation}

\begin{table}[!t]
\fontsize{8pt}{8pt}\selectfont
\setlength{\tabcolsep}{3pt}
\centering
\caption{Comparison of finetuning  IN-F5 with different hours per language (L). S - Seen speakers, U - Unseen speakers, M - Monoglot, P - Polyglot}
\label{tab:scaling-effects}
\begin{tabular}{@{}rccccc@{}}
\toprule
\textbf{Setup} & \textbf{MUSHRA} & \textbf{SSIM-S} & \textbf{SSIM-U} & \textbf{WER-M(\%)} & \textbf{WER-P(\%)} \\ \midrule
100h / L       & \textbf{64.3}   & \textbf{93.1}               & 93.9                 & 32.6              & \textbf{30.5}     \\
10h / L        & 61.5            & \textbf{93.1}               & \textbf{94.0}                 & \textbf{31.3}     & 31.8              \\
1h / L         & 33.7            & 92.7               & 93.5                 & 59.4              & 60.4              \\ \bottomrule
\end{tabular}
\end{table}

\noindent In Table \ref{tab:scaling-effects}, we assess emergence across data scales, in terms of (i) naturalness (ii) speaker similarity and (iii) intelligibility. The scores reveal that fine-tuning on just 10 hours per language (10h/L) retains key emergent behaviors compared to the larger fine-tuning, with an average reduction of 0.8\% in performance across metrics. While scaling data definitely shows improvements, in low-resource settings one may be forced to train with lesser data. We realize that training for longer can still help the data-poor setup, with a model fine-tuned on 10 hours for 150 K steps achieving a WER of 31.3\% compared to fine-tuning on 100 hours for 120K steps achieving a WER of 40.0\% in a monoglot setup. Finally, we note that English pre-training severely benefits all models, even the lowest resource setup (1h/L), in achieving high speaker similarity scores ($> 92$\%) for seen and unseen speakers. This suggests that pretraining on English may perhaps benefit models in implicitly learning speaker characteristics that transfer well during  multilingual adaptation.



\subsection{Building Zero-Resource TTS}

IN-F5 excels at zero-shot cross-lingual TTS. To demonstrate this, we evaluate two scenarios: a simulated zero-resource setting for Bhojpuri and a truly zero-resource setup for Tulu. We use IN-F5 to synthesize Bhojpuri speech leveraging a pleasant and professional voice from Rasa Maithili. Notably, IN-F5 has never seen Bhojpuri or Maithili in training, making this a strong test of its cross-lingual transfer ability. We intentionally select the expressive speaker from a related language (Maithili), as it can lead to a prompt-TTS generating samples more natural than a less expressive ground-truth recording. IN-F5 achieves a MUSHRA score of 82 and surpasses human recordings (67.1) in the simulated zero-resource setting. In Tulu, the model scores highly indicating intelligible and natural synthesis. Next, we task a language expert to validate 1 hour of synthesized data for both Bhojpuri and Tulu. Using the accepted samples, we self-train (ST) IN-F5 and release the validated dataset to encourage further TTS research for low-resource languages. Table \ref{tab:zero-resource} shows that self-training improves Bhojpuri slightly but has no positive impact on Tulu, which already scores highly.We humbly request readers to interpret MUSHRA scores with caution, as each language is evaluated by a single language expert, due to the challenge of sourcing reliable listeners for low-resource languages. Despite this, our results affirm IN-F5’s ability to generate highly convincing TTS for zero-resource languages, offering a promising path toward scalable multilingual speech synthesis in India.

\begin{table}[!t]
\fontsize{8pt}{8pt}\selectfont
\setlength{\tabcolsep}{4pt}
\centering
\caption{MUSHRA scores for Zero-Resource TTS, before and after self-training (ST), compared against human recordings.}
\label{tab:zero-resource}

\begin{tabular}{@{}lccc@{}}
\toprule
\textbf{}  & \textbf{Bhojpuri}    & \textbf{Tulu}         \\ \midrule
IN-F5      & 82.0  ± 6.2                            & \textbf{93.62  ± 4.2} \\
IN-F5 + ST & \textbf{83.9  ±  6.2} & 93.12  ± 5.8          \\
Human      & 67.1  ±  3.5      & -                     \\ \bottomrule
\end{tabular}
\end{table}

\subsection{Advancing State-Of-The-Art for Indian Languages}
We benchmark IN-F5 against prior state-of-the-art models on 8 out of the 11 Indian languages included in the official Rasa \cite{ai4bharat2024rasa} test set. The MUSHRA scores, presented in Table \ref{tab:baselines}, highlight a significant 8-point improvement over VoiceCraft and a 14-point gain over FastSpeech2-HS, establishing IN-F5 as a new topline for Indian TTS. Beyond these numerical gains, IN-F5 is the only model to enter the "Excellent" quality range (MUSHRA 80-100), a milestone never before achieved in Indian TTS.

\begin{table}[!t]
\centering
\fontsize{8pt}{8pt}\selectfont
\setlength{\tabcolsep}{2pt}
\caption{Comparison of IN-F5 against prior SOTA Indian TTS.}
\label{tab:baselines}
\begin{tabular}{@{}lccc@{}}
\toprule
\textbf{Systems} & \multicolumn{1}{l}{\textbf{MUSHRA}} & \multicolumn{1}{l}{\textbf{WER}} & \multicolumn{1}{l}{\textbf{S-SIM}} \\ \midrule
FastPitch~\cite{Kumar2022TowardsBT}      & 63.8 ± 2.2          & \textbf{18.0} & 89.7          \\
FastSpeech2-HS~\cite{DBLP:conf/asru/PrakashUM23} & 66.3 ± 2.2        & 27.2          & 90.1          \\
VoiceCraft~\cite{DBLP:conf/acl/Peng00MH24}     & 73.0 ± 1.8         & 21.0          & 95.7          \\
\textbf{IN-F5} & \textbf{80.5 ± 1.5} & 19.2          & \textbf{97.3} \\ 
Human          & 91.8 ± 0.7         & 18.4          & 97.0          \\ \bottomrule
\end{tabular}
\end{table}

\section{Conclusion}
We establish IN-F5 as a state-of-the-art TTS system for Indian languages, demonstrating that fine-tuning a large-scale English model enables high-quality speech synthesis while unlocking polyglot fluency, voice cloning, and code-mixing. Contrary to expectations, 
fine-tuning without English yields the most natural adaptation, 
We also realize zero-resource TTS for Bhojpuri and Tulu speech via transfer learning and human-in-the-loop refinement. Our approach offers a practical, scalable solution for expanding TTS to underrepresented languages, bridging the gap toward inclusive and human-like speech synthesis in India.



\bibliographystyle{IEEEtran}
\bibliography{mybib}

\begin{thebibliography}{10}
\providecommand{\url}[1]{#1}
\csname url@samestyle\endcsname
\providecommand{\newblock}{\relax}
\providecommand{\bibinfo}[2]{#2}
\providecommand{\BIBentrySTDinterwordspacing}{\spaceskip=0pt\relax}
\providecommand{\BIBentryALTinterwordstretchfactor}{4}
\providecommand{\BIBentryALTinterwordspacing}{\spaceskip=\fontdimen2\font plus
\BIBentryALTinterwordstretchfactor\fontdimen3\font minus \fontdimen4\font\relax}
\providecommand{\BIBforeignlanguage}[2]{{%
\expandafter\ifx\csname l@#1\endcsname\relax
\typeout{** WARNING: IEEEtran.bst: No hyphenation pattern has been}%
\typeout{** loaded for the language `#1'. Using the pattern for}%
\typeout{** the default language instead.}%
\else
\language=\csname l@#1\endcsname
\fi
#2}}
\providecommand{\BIBdecl}{\relax}
\BIBdecl

\bibitem{DBLP:journals/corr/abs-2409-00750}
\BIBentryALTinterwordspacing
Y.~Wang, H.~Zhan, L.~Liu, R.~Zeng, H.~Guo, J.~Zheng, Q.~Zhang, S.~Zhang, and Z.~Wu, ``Maskgct: Zero-shot text-to-speech with masked generative codec transformer,'' \emph{CoRR}, vol. abs/2409.00750, 2024. [Online]. Available: \url{https://doi.org/10.48550/arXiv.2409.00750}
\BIBentrySTDinterwordspacing

\bibitem{DBLP:journals/pami/TanCLCZLWLYHZQSL24}
\BIBentryALTinterwordspacing
X.~Tan, J.~Chen, H.~Liu, J.~Cong, C.~Zhang, Y.~Liu, X.~Wang, Y.~Leng, Y.~Yi, L.~He, S.~Zhao, T.~Qin, F.~K. Soong, and T.~Liu, ``Naturalspeech: End-to-end text-to-speech synthesis with human-level quality,'' \emph{{IEEE} Trans. Pattern Anal. Mach. Intell.}, vol.~46, no.~6, pp. 4234--4245, 2024. [Online]. Available: \url{https://doi.org/10.1109/TPAMI.2024.3356232}
\BIBentrySTDinterwordspacing

\bibitem{anastassiou2024seed}
P.~Anastassiou, J.~Chen, J.~Chen, Y.~Chen, Z.~Chen, Z.~Chen, J.~Cong, L.~Deng, C.~Ding, L.~Gao \emph{et~al.}, ``Seed-tts: A family of high-quality versatile speech generation models,'' \emph{arXiv preprint arXiv:2406.02430}, 2024.

\bibitem{shen2024naturalspeech}
\BIBentryALTinterwordspacing
K.~Shen, Z.~Ju, X.~Tan, E.~Liu, Y.~Leng, L.~He, T.~Qin, sheng zhao, and J.~Bian, ``Naturalspeech 2: Latent diffusion models are natural and zero-shot speech and singing synthesizers,'' in \emph{The Twelfth International Conference on Learning Representations}, 2024. [Online]. Available: \url{https://openreview.net/forum?id=Rc7dAwVL3v}
\BIBentrySTDinterwordspacing

\bibitem{DBLP:conf/acl/Peng00MH24}
\BIBentryALTinterwordspacing
P.~Peng, P.~Huang, S.~Li, A.~Mohamed, and D.~Harwath, ``Voicecraft: Zero-shot speech editing and text-to-speech in the wild,'' in \emph{Proceedings of the 62nd Annual Meeting of the Association for Computational Linguistics (Volume 1: Long Papers), {ACL} 2024, Bangkok, Thailand, August 11-16, 2024}, L.~Ku, A.~Martins, and V.~Srikumar, Eds.\hskip 1em plus 0.5em minus 0.4em\relax Association for Computational Linguistics, 2024, pp. 12\,442--12\,462. [Online]. Available: \url{https://doi.org/10.18653/v1/2024.acl-long.673}
\BIBentrySTDinterwordspacing

\bibitem{borsos2023soundstorm}
Z.~Borsos, M.~Sharifi, D.~Vincent, E.~Kharitonov, N.~Zeghidour, and M.~Tagliasacchi, ``Soundstorm: Efficient parallel audio generation,'' \emph{arXiv preprint arXiv:2305.09636}, 2023.

\bibitem{ju2024naturalspeech}
Z.~Ju, Y.~Wang, K.~Shen, X.~Tan, D.~Xin, D.~Yang, Y.~Liu, Y.~Leng, K.~Song, S.~Tang, Z.~Wu, T.~Qin, X.-Y. Li, W.~Ye, S.~Zhang, J.~Bian, L.~He, J.~Li, and S.~Zhao, ``Naturalspeech 3: Zero-shot speech synthesis with factorized codec and diffusion models,'' 2024.

\bibitem{le2024voicebox}
M.~Le, A.~Vyas, B.~Shi, B.~Karrer, L.~Sari, R.~Moritz, M.~Williamson, V.~Manohar, Y.~Adi, J.~Mahadeokar \emph{et~al.}, ``Voicebox: Text-guided multilingual universal speech generation at scale,'' \emph{Advances in neural information processing systems}, vol.~36, 2024.

\bibitem{du2024cosyvoice}
Z.~Du, Y.~Wang, Q.~Chen, X.~Shi, X.~Lv, T.~Zhao, Z.~Gao, Y.~Yang, C.~Gao, H.~Wang \emph{et~al.}, ``Cosyvoice 2: Scalable streaming speech synthesis with large language models,'' \emph{arXiv preprint arXiv:2412.10117}, 2024.

\bibitem{chen2024f5}
Y.~Chen, Z.~Niu, Z.~Ma, K.~Deng, C.~Wang, J.~Zhao, K.~Yu, and X.~Chen, ``F5-tts: A fairytaler that fakes fluent and faithful speech with flow matching,'' \emph{arXiv preprint arXiv:2410.06885}, 2024.

\bibitem{casanova2022yourtts}
E.~Casanova, J.~Weber, C.~D. Shulby, A.~C. Junior, E.~G{\"o}lge, and M.~A. Ponti, ``Yourtts: Towards zero-shot multi-speaker tts and zero-shot voice conversion for everyone,'' in \emph{International Conference on Machine Learning}.\hskip 1em plus 0.5em minus 0.4em\relax PMLR, 2022, pp. 2709--2720.

\bibitem{Kumar2022TowardsBT}
\BIBentryALTinterwordspacing
G.~K. Kumar, V.~PraveenS., P.~Kumar, M.~M. Khapra, and K.~Nandakumar, ``Towards building text-to-speech systems for the next billion users,'' \emph{ICASSP 2023 - 2023 IEEE International Conference on Acoustics, Speech and Signal Processing (ICASSP)}, pp. 1--5, 2022. [Online]. Available: \url{https://api.semanticscholar.org/CorpusID:253581696}
\BIBentrySTDinterwordspacing

\bibitem{varadhan2024rethinking}
P.~S. Varadhan, A.~Gulati, A.~Sankar, S.~Anand, A.~Gupta, A.~Mukherjee, S.~K. Marepally, A.~Bhatia, S.~Jaju, S.~Bhooshan \emph{et~al.}, ``Rethinking mushra: Addressing modern challenges in text-to-speech evaluation,'' \emph{arXiv preprint arXiv:2411.12719}, 2024.

\bibitem{anand24_interspeech}
S.~Anand, P.~{Srinivasa Varadhan}, A.~Sankar, G.~Raju, and M.~M. Khapra, ``Enhancing out-of-vocabulary performance of indian tts systems for practical applications through low-effort data strategies,'' in \emph{Interspeech 2024}, 2024, pp. 1200--1204.

\bibitem{he2024emilia}
H.~He, Z.~Shang, C.~Wang, X.~Li, Y.~Gu, H.~Hua, L.~Liu, C.~Yang, J.~Li, P.~Shi \emph{et~al.}, ``Emilia: An extensive, multilingual, and diverse speech dataset for large-scale speech generation,'' in \emph{2024 IEEE Spoken Language Technology Workshop (SLT)}.\hskip 1em plus 0.5em minus 0.4em\relax IEEE, 2024, pp. 885--890.

\bibitem{sankar2024indicvoices-r}
\BIBentryALTinterwordspacing
A.~Sankar, S.~Anand, P.~S. Varadhan, S.~Thomas, M.~Singal, S.~Kumar, D.~Mehendale, A.~Krishana, G.~Raju, and M.~M. Khapra, ``Indicvoices-r: Unlocking a massive multilingual multi-speaker speech corpus for scaling indian {TTS},'' \emph{NeurIPS 2024 Datasets and Benchmarks track, Vancouver, Canada}, vol. abs/2409.05356, 2024. [Online]. Available: \url{https://doi.org/10.48550/arXiv.2409.05356}
\BIBentrySTDinterwordspacing

\bibitem{iiith2024limmits}
A.~Singh, A.~Nagireddi, D.~G, J.~Bandekar, R.~R, S.~Badiger, S.~Udupa, P.~K. Ghosh, H.~A. Murthy, P.~Kumar, K.~Tokuda, M.~Hasegawa-Johnson, and P.~Olbrich, ``Limmits’24: Multi-speaker, multi-lingual indic tts with voice cloning,'' in \emph{2024 IEEE International Conference on Acoustics, Speech, and Signal Processing Workshops (ICASSPW)}, 2024, pp. 61--62.

\bibitem{narayanan-aepli-2024-tulu}
\BIBentryALTinterwordspacing
M.~Narayanan and N.~Aepli, ``A {T}ulu resource for machine translation,'' in \emph{Proceedings of the 2024 Joint International Conference on Computational Linguistics, Language Resources and Evaluation (LREC-COLING 2024)}, N.~Calzolari, M.-Y. Kan, V.~Hoste, A.~Lenci, S.~Sakti, and N.~Xue, Eds.\hskip 1em plus 0.5em minus 0.4em\relax Torino, Italia: ELRA and ICCL, May 2024, pp. 1756--1767. [Online]. Available: \url{https://aclanthology.org/2024.lrec-main.155/}
\BIBentrySTDinterwordspacing

\bibitem{ai4bharat2024rasa}
P.~S. Varadhan, A.~Sankar, G.~Raju, and M.~M. Khapra, ``{Rasa: Building Expressive Speech Synthesis Systems for Indian Languages in Low-resource Settings},'' in \emph{Proc. INTERSPEECH 2024}, 2024.

\bibitem{baby2016resources}
A.~Baby, A.~L. Thomas, N.~Nishanthi, T.~Consortium \emph{et~al.}, ``Resources for indian languages,'' in \emph{Proceedings of Text, Speech and Dialogue}, 2016.

\bibitem{abraham2020crowdsourcing}
B.~Abraham, D.~Goel, D.~Siddarth, K.~Bali, M.~Chopra, M.~Choudhury, P.~Joshi, P.~Jyoti, S.~Sitaram, and V.~Seshadri, ``Crowdsourcing speech data for low-resource languages from low-income workers,'' in \emph{Proceedings of the 12th Language Resources and Evaluation Conference}, 2020, pp. 2819--2826.

\bibitem{javed2024indicvoices}
T.~Javed, J.~A. Nawale, E.~I. George, S.~Joshi, K.~S. Bhogale, D.~Mehendale, I.~V. Sethi, A.~Ananthanarayanan, H.~Faquih, P.~Palit \emph{et~al.}, ``Indicvoices: Towards building an inclusive multilingual speech dataset for indian languages,'' \emph{arXiv preprint arXiv:2403.01926}, 2024.

\bibitem{DBLP:journals/jstsp/ChenWCWLCLKYXWZ22}
\BIBentryALTinterwordspacing
S.~Chen, C.~Wang, Z.~Chen, Y.~Wu, S.~Liu, Z.~Chen, J.~Li, N.~Kanda, T.~Yoshioka, X.~Xiao, J.~Wu, L.~Zhou, S.~Ren, Y.~Qian, Y.~Qian, J.~Wu, M.~Zeng, X.~Yu, and F.~Wei, ``Wavlm: Large-scale self-supervised pre-training for full stack speech processing,'' \emph{{IEEE} J. Sel. Top. Signal Process.}, vol.~16, no.~6, pp. 1505--1518, 2022. [Online]. Available: \url{https://doi.org/10.1109/JSTSP.2022.3188113}
\BIBentrySTDinterwordspacing

\bibitem{DBLP:conf/asru/PrakashUM23}
\BIBentryALTinterwordspacing
A.~Prakash, S.~Umesh, and H.~A. Murthy, ``Towards developing state-of-the-art {TTS} synthesisers for 13 indian languages with signal processing aided alignments,'' in \emph{{IEEE} Automatic Speech Recognition and Understanding Workshop, {ASRU} 2023, Taipei, Taiwan, December 16-20, 2023}.\hskip 1em plus 0.5em minus 0.4em\relax {IEEE}, 2023, pp. 1--8. [Online]. Available: \url{https://doi.org/10.1109/ASRU57964.2023.10389630}
\BIBentrySTDinterwordspacing

\end{thebibliography}

\end{document}